# Pac-learning Recursive Logic Programs: Negative Results

**William W. Cohen**                                    WCOHEN@RESEARCH.ATT.COM
*AT&T Bell Laboratories*
*600 Mountain Avenue, Murray Hill, NJ 07974 USA*

## Abstract

In a companion paper it was shown that the class of constant-depth determinate $k$-ary recursive clauses is efficiently learnable. In this paper we present negative results showing that any natural generalization of this class is hard to learn in Valiant's model of pac-learnability. In particular, we show that the following program classes are cryptographically hard to learn: programs with an unbounded number of constant-depth linear recursive clauses; programs with one constant-depth determinate clause containing an unbounded number of recursive calls; and programs with one linear recursive clause of constant locality. These results immediately imply the non-learnability of any more general class of programs. We also show that learning a constant-depth determinate program with either two linear recursive clauses or one linear recursive clause and one non-recursive clause is as hard as learning boolean DNF. Together with positive results from the companion paper, these negative results establish a boundary of efficient learnability for recursive function-free clauses.

## 1. Introduction

*Inductive logic programming (ILP)* (Muggleton, 1992; Muggleton & De Raedt, 1994) is an active area of machine learning research in which the hypotheses of a learning system are expressed in a logic programming language. While many different learning problems have been considered in ILP, including some of great practical interest (Muggleton, King, & Sternberg, 1992; King, Muggleton, Lewis, & Sternberg, 1992; Zelle & Mooney, 1994; Cohen, 1994b), a class of problems that is frequently considered is to reconstruct simple list-processing or arithmetic functions from examples. A prototypical problem of this sort might be learning to append two lists. Often, this sort of task is attempted using only randomly-selected positive and negative examples of the target concept.

Based on its similarity to the problems studied in the field of automatic programming from examples (Summers, 1977; Biermann, 1978), we will (informally) call this class of learning tasks *automatic logic programming* problems. While a number of experimental systems have been built (Quinlan & Cameron-Jones, 1993; Aha, Lapointe, Ling, & Matwin, 1994), the experimental success in automatic logic programming systems has been limited. One common property of automatic logic programming problems is the presence of *recursion*. The goal of this paper is to explore by analytic methods the computational limitations on learning recursive programs in Valiant's model of pac-learnability (1984). (In brief, this model requires that an accurate approximation of the target concept be found in polynomial time using a polynomial-sized set of labeled examples, which are chosen stochastically.) While it will surprise nobody that such limitations exist, it is far from obvious from previous research where these limits lie: there are few provably fast methods for learning recursive logic programs, and even fewer meaningful negative results.

The starting point for this investigation is a series of positive learnability results appearing in a companion paper (Cohen, 1995). These results show that a single constant-depth determinate clause with a constant number of "closed" recursive calls is pac-learnable. They also show that a two-clause constant-depth determinate program consisting of one nonrecursive clause and one recursive clause of the type described above is pac-learnable, if some additional "hints" about the target concept are provided.

In this paper, we analyze a number of generalizations of these learnable languages. We show that that relaxing any of the restrictions leads to difficult learning problems: in particular, learning problems that are either as hard as learning DNF (an open problem in computational learning theory), or as hard as cracking certain presumably secure cryptographic schemes. The main contribution of this paper, therefore, is a delineation of the boundaries of learnability for recursive logic programs.

The paper is organized as follows. In Section 2 we define the classes of logic programs and the learnability models that are used in this paper. In Section 3 we present cryptographic hardness results for two classes of constant-depth determinate recursive programs: programs with $n$ linear recursive clauses, and programs with one $n$-ary recursive clause. We also analyze the learnability of clauses of constant locality, another class of clauses that is pac-learnable in the nonrecursive case, and show that even a single linearly recursive local clause is cryptographically hard to learn. We then turn, in Section 4, to the analysis of even more restricted classes of recursive programs. We show that two different classes of constant-depth determinate programs are prediction-equivalent to boolean DNF: the class of programs containing a single linear recursive clause and a single nonrecursive clause, and the class of programs containing two linearly recursive clauses. Finally, we summarize the results of this paper and its companion, discuss related work, and conclude.

Although this paper can be read independently of its companion paper we suggest that readers planning to read both papers begin with the companion paper (Cohen, 1995).

## 2. Background

For completeness, we will now present the technical background needed to state our results; however, aside from Sections 2.2 and 2.3, which introduce polynomial predictability and prediction-preserving reducibilities, respectively, this background closely follows that presented in the companion paper (Cohen, 1995). Readers are encouraged to skip this section if they are already familiar with the material.

### 2.1 Logic Programs

We will assume that the reader has some familiarity in logic programming (such as can be obtained by reading one of the standard texts (Lloyd, 1987).) Our treatment of logic programs differs only in that we will usually consider the body of a clause to be an ordered set of literals. We will also consider only logic programs without function symbols—*i.e.*, programs written in Datalog.

The semantics of a Datalog program $P$ will be defined relative to to a *database*, $DB$, which is a set of ground atomic facts. (When convenient, we will also think of $DB$ as a conjunction of ground unit clauses). In particular, we will interpret $P$ and $DB$ as a subset of the set of all *extended instances*. An *extended instance* is a pair $(f, D)$ in which the *instance fact* $f$ is a ground fact, and the *description* $D$ is a set of ground unit clauses. An extended instance $(f, D)$ is covered by $(P, DB)$ iff

$$DB \wedge D \wedge P \vdash f$$

If extended instances are allowed, then function-free programs can encode many computations that are usually represented with function symbols. For example, a function-free program that tests to see if a list is the *append* of two other lists can be written as follows:

**Program $P$:**
append(Xs,Ys,Ys) ←
    null(Xs).
append(Xs,Ys,Zs) ←
    components(Xs,X,Xs1) ∧
    components(Zs,X,Zs1) ∧
    append(Xs1,Ys,Zs1).

**Database $DB$:**
null(nil).

Here the predicate *components(A,B,C)* means that $A$ is a list with head $B$ and tail $C$; thus an extended instance equivalent to *append([1,2],[3],[1,2,3])* would have the instance fact $f = append(list12, list3, list123)$ and a description containing these atoms:

components(list12,1,list2), components(list2,2,nil),
components(list123,1,list23), components(list23,2,list3),
components(list3,3,nil)

The use of extended instances and function-free programs is closely related to "flattening" (Rouveirol, 1994; De Raedt & Džeroski, 1994); some experimental learning systems also impose a similar restriction (Quinlan, 1990; Pazzani & Kibler, 1992). Another motivation for using extended instances is technical. Under the (sometimes quite severe) syntactic restrictions considered in this paper, there are often only a polynomial number of possible ground facts—*i.e.*, the Herbrand base is polynomial. Hence if programs were interpreted in the usual model-theoretic way it would be possible to learn a program equivalent to any given target by simply memorizing the appropriate subset of the Herbrand base. However, if programs are interpreted as sets of extended instances, such trivial learning algorithms become impossible; even for extremely restricted program classes there are still an exponential number of extended instances of size $n$. Further discussion can be found in the companion paper (Cohen, 1995).

Below we will define some of the terminology for logic programs that will be used in this paper.

### 2.1.1 INPUT/OUTPUT VARIABLES

If $A \leftarrow B_1 \wedge \ldots \wedge B_r$ is an (ordered) definite clause, then the *input variables* of the literal $B_i$ are those variables which also appear in the clause $A \leftarrow B_1 \wedge \ldots \wedge B_{i-1}$; all other variables appearing in $B_i$ are called *output variables*.

### 2.1.2 Types of Recursion

A literal in the body of a clause is a *recursive literal* if it has the same predicate symbol and arity as the head of the clause. If every clause in a program has at most one recursive literal, the program is *linear recursive*. If every clause in a program has at most $k$ recursive literals, the program is *$k$-ary recursive*. If every recursive literal in a program contains no output variables, the program is *closed recursive*.

### 2.1.3 Depth

The *depth* of a variable appearing in a (ordered) clause $A \leftarrow B_1 \land \ldots \land B_r$ is defined as follows. Variables appearing in the head of a clause have depth zero. Otherwise, let $B_i$ be the first literal containing the variable $V$, and let $d$ be the maximal depth of the input variables of $B_i$; then the depth of $V$ is $d+1$. The depth of a clause is the maximal depth of any variable in the clause.

### 2.1.4 Determinacy

The literal $B_i$ in the clause $A \leftarrow B_1 \land \ldots \land B_r$ is *determinate* iff for every possible substitution $\sigma$ that unifies $A$ with some fact $e$ such that

$$DB \vdash B_1\sigma \land \ldots \land B_{i-1}\sigma$$

there is at most one maximal substitution $\theta$ so that $DB \vdash B_i\sigma\theta$. A clause is *determinate* if all of its literals are determinate. Informally, determinate clauses are those that can be evaluated without backtracking by a Prolog interpreter.

The term *$ij$-determinate* (Muggleton & Feng, 1992) is sometimes used for programs that are depth $i$, determinate, and contain literals of arity $j$ or less. A number of experimental systems exploit restrictions associated with limited depth and determinacy (Muggleton & Feng, 1992; Quinlan, 1991; Lavrač & Džeroski, 1992; Cohen, 1993c). The learnability of constant-depth determinate clauses has also received some formal study (Džeroski, Muggleton, & Russell, 1992; Cohen, 1993a).

### 2.1.5 Mode Constraints and Declarations

*Mode declarations* are commonly used in analyzing Prolog code or describing Prolog code; for instance, the mode declaration "*components*$(+, -, -)$" indicates that the predicate *components* can be used when its first argument is an input and its second and third arguments are outputs. Formally, we define the *mode* of a literal $L$ appearing in a clause $C$ to be a string $s$ such that the initial character of $s$ is the predicate symbol of $L$, and for $j > 1$ the $j$-th character of $s$ is a "+" if the $(j-1)$-th argument of $L$ is an input variable and a "−" if the $(j-1)$-th argument of $L$ is an output variable. (This definition assumes that all arguments to the head of a clause are inputs; this is justified since we are considering only how clauses behave in classifying extended instances, which are ground.) A *mode constraint* is a set of mode strings $R = \{s_1, \ldots, s_k\}$, and a clause $C$ is said to *satisfy* a mode constraint $R$ for $p$ if for every literal $L$ in the body of $C$, the mode of $L$ is in $R$.

We define a *declaration* to be a tuple $(p, a', R)$ where $p$ is a predicate symbol, $a'$ is an integer, and $R$ is a mode constraint. We will say that a clause $C$ *satisfies* a declaration if the head of $C$ has arity $a'$ and predicate symbol $p$, and if for every literal $L$ in the body of $C$ the mode of $L$ appears in $R$.

### 2.1.6 DETERMINATE MODES

In a typical setting, that facts in the database $DB$ and extended instances are not arbitrary: instead, they are representative of some "real" predicate, which may obey certain restrictions. Let us assume that all database and extended-instance facts will be drawn from some (possibly infinite) set $\mathcal{F}$. Informally, a mode is determinate if the input positions of the facts in $\mathcal{F}$ functionally determine the output positions. Formally, if $f = p(t_1, \ldots, t_k)$ is a fact with predicate symbol $p$ and $p\alpha$ is a mode, then define $inputs(f, p\alpha)$ to be $\langle t_{i_1}, \ldots, t_{i_k} \rangle$, where $i_1, \ldots, i_k$ are the indices of $\alpha$ containing a "+", and define $outputs(f, p\alpha)$ to be $\langle t_{j_1}, \ldots, t_{j_l} \rangle$, where $j_1, \ldots, j_l$ are the indices of $\alpha$ containing a "−". We define a mode string $p\alpha$ for a predicate $p$ to be *determinate for $\mathcal{F}$* iff

$$\{\langle inputs(f, p\alpha), outputs(f, p\alpha) \rangle : f \in \mathcal{F}\}$$

is a function. Any clause that satisfies a declaration $Dec \in \mathcal{D}et\mathcal{DEC}$ must be determinate.

The set of all declarations containing only modes determinate for $\mathcal{F}$ will be denoted $\mathcal{D}et\mathcal{DEC}_{\mathcal{F}}$. Since in this paper the set $\mathcal{F}$ will be assumed to be fixed, we will generally omit the subscript.

### 2.1.7 BOUNDS ON PREDICATE ARITY

We will use the notation $a$-$\mathcal{DB}$ for the set of all databases that contain only facts of arity $a$ or less, and $a$-$\mathcal{DEC}$ for the set of all declarations $(p, a', R)$ such that every string $s \in R$ is of length $a + 1$ or less.

### 2.1.8 SIZE MEASURES

The learning models presented in the following section will require the learner to use resources polynomial in the size of its inputs. Assuming that all predicates are arity $a$ or less for some constant $a$ allows very simple size measures to be used. In this paper, we will measure the size of a database $DB$ by its cardinality; the size of an extended instance $(f, D)$ by the cardinality of $D$; the size of a declaration $(p, a', R)$ by the cardinality of $R$; and the size of a clause $A \leftarrow B_1 \wedge \ldots \wedge B_r$ by the number of literals in its body.

## 2.2 A Model of Learnability

### 2.2.1 PRELIMINARIES

Let $X$ be a set. We will call $X$ the *domain*, and call the elements of $X$ *instances*. Define a *concept* $C$ over $X$ to be a representation of some subset of $X$, and define a *language* LANG to be a set of concepts. In this paper, we will be rather casual about the distinction between a concept and the set it represents; when there is a risk of confusion we will refer to the set represented by a concept $C$ as the *extension of $C$*. Two sets $C_1$ and $C_2$ with the same extension are said to be *equivalent*. Define an *example of $C$* to be a pair $(e, b)$ where $b = 1$ if $e \in C$ and $b = 0$ otherwise. If $D$ is a probability distribution function, a *sample of $C$ from* $X$ *drawn according to* $D$ is a pair of multisets $S^+, S^-$ drawn from the domain $X$ according to $D$, $S^+$ containing only positive examples of $C$, and $S^-$ containing only negative ones.

Associated with $X$ and LANG are two *size complexity measures*, for which we will use the following notation:

- The size complexity of a concept $C \in$ LANG is written $\|C\|$.

- The size complexity of an instance $e \in X$ is written $\|e\|$.

- If $S$ is a set, $S_n$ stands for the set of all elements of $S$ of size complexity no greater than $n$. For instance, $X_n = \{e \in X : \|e\| \leq n\}$ and LANG$_n = \{C \in$ LANG $: \|C\| \leq n\}$.

We will assume that all size measures are polynomially related to the number of bits needed to represent $C$ or $e$; this holds, for example, for the size measures for logic programs and databases defined above.

### 2.2.2 POLYNOMIAL PREDICTABILITY

We now define polynomial predictability as follows. A language LANG is *polynomially predictable* iff there is an algorithm PACPREDICT and a polynomial function $m(\frac{1}{\epsilon}, \frac{1}{\delta}, n_e, n_t)$ so that for every $n_t > 0$, every $n_e > 0$, every $C \in$ LANG$_{n_t}$, every $\epsilon : 0 < \epsilon < 1$, every $\delta : 0 < \delta < 1$, and every probability distribution function $D$, PACPREDICT has the following behavior:

1. given a sample $S^+, S^-$ of $C$ from $X_{n_e}$ drawn according to $D$ and containing at least $m(\frac{1}{\epsilon}, \frac{1}{\delta}, n_e, n_t)$ examples, PACPREDICT outputs a hypothesis $H$ such that

$$Prob(D(H - C) + D(C - H) > \epsilon) < \delta$$

where the probability is taken over the possible samples $S^+$ and $S^-$ and (if PACPREDICT is a randomized algorithm) over any coin flips made by PACPREDICT;

2. PACPREDICT runs in time polynomial in $\frac{1}{\epsilon}$, $\frac{1}{\delta}$, $n_e$, $n_t$, and the number of examples; and

3. The hypothesis $H$ can be evaluated in polynomial time.

The algorithm PACPREDICT is called a *prediction algorithm* for LANG, and the function $m(\frac{1}{\epsilon}, \frac{1}{\delta}, n_e, n_t)$ is called the *sample complexity* of PACPREDICT. We will sometimes abbreviate "polynomial predictability" as "predictability".

The first condition in the definition merely states that the error rate of the hypothesis must (usually) be low, as measured against the probability distribution $D$ from which the training examples were drawn. The second condition, together with the stipulation that the sample size is polynomial, ensures that the total running time of the learner is polynomial. The final condition simply requires that the hypothesis be usable in the very weak sense that it can be used to make predictions in polynomial time. Notice that this is a worst case learning model, as the definition allows an adversarial choice of all the inputs of the learner.

### 2.2.3 RELATION TO OTHER MODELS

The model of polynomial predictability has been well-studied (Pitt & Warmuth, 1990), and is a weaker version of Valiant's (1984) criterion of *pac-learnability*. A language LANG is *pac-learnable* iff there is an algorithm PACLEARN so that

1. PACLEARN satisfies all the requirements in the definition of polynomial predictability, and

2. on inputs $S^+$ and $S^-$, PACLEARN always outputs a hypothesis $H \in$ LANG.

Thus if a language is pac-learnable it is predictable.

In the companion paper (Cohen, 1995), our positive results are all expressed in the model of identifiability from equivalence queries, which is strictly stronger than pac-learnability; that is, anything that is learnable from equivalence queries is also necessarily pac-learnable.[1] Since this paper contains only negative results, we will use the the relatively weak model of predictability. Negative results in this model immediately translate to negative results in the stronger models; if a language is not predictable, it cannot be pac-learnable, nor identifiable from equivalence queries.

### 2.2.4 BACKGROUND KNOWLEDGE IN LEARNING

In a typical ILP system, the setting is slightly different, as the user usually provides clues about the target concept in addition to the examples, in the form of a database $DB$ of "background knowledge" and a set of declarations. To account for these additional inputs it is necessary to extend the framework described above to a setting where the learner accepts inputs other than training examples. Following the formalization used in the companion paper (Cohen, 1995), we will adopt the notion of a "language family".

If LANG is a set of clauses, $DB$ is a database and $Dec$ is a declaration, we will define LANG$[DB, Dec]$ to be the set of all pairs $(C, DB)$ such that $C \in$ LANG and $C$ satisfies $Dec$. Semantically, such a pair will denote the set of all extended instances $(f, D)$ covered by $(C, DB)$. Next, if $\mathcal{DB}$ is a set of databases and $\mathcal{DEC}$ is a set of declarations, then define

$$\text{LANG}[\mathcal{DB}, \mathcal{DEC}] = \{\text{LANG}[DB, Dec] : DB \in \mathcal{DB} \text{ and } Dec \in \mathcal{DEC}\}$$

This set of languages is called a *language family*.

We will now extend the definition of predictability queries to language families as follows. A language family LANG$[\mathcal{DB}, \mathcal{DEC}]$ is polynomially predictable iff every language in the set is predictable. A language family LANG$[\mathcal{DB}, \mathcal{DEC}]$ is *polynomially predictable* iff there is a single algorithm IDENTIFY$(DB, Dec)$ that predicts every LANG$[DB, Dec]$ in the family given $DB$ and $Dec$.

The usual model of polynomial predictability is worst-case over all choices of the target concept and the distribution of examples. The notion of polynomial predictability of a language family extends this model in the natural way; the extended model is also worst-case over all possible choices for database $DB \in \mathcal{DB}$ and $Dec \in \mathcal{DEC}$. This worst-case model may seem unintuitive, since one typically assumes that the database $DB$ is provided by a helpful user, rather than an adversary. However, the worst-case model is reasonable because learning is allowed to take time polynomial in the size of smallest target concept *in the set* Lang$[DB, Dec]$; this means that if the database given by the user is such that the target concept cannot be encoded succinctly (or at all) learning is allowed to take more time.

---

1. An equivalence query is a question of the form "is $H$ equivalent to the target concept?" which is answered with either "yes" or a counterexample. Identification by equivalence queries essentially means that the target concept can be exactly identified in polynomial time using a polynomial of such queries.

Notice that for a language family Lang$[DB, Dec]$ to be polynomially predictable, every language in the family must be polynomially predictable. Thus to show that a family is not polynomially predictable it is sufficient to construct one language in the family for which learning is hard. The proofs of this paper will all have this form.

## 2.3 Prediction-Preserving Reducibilities

The principle technical tool used in our negative results in the notion of *prediction-preserving reducibility*, as introduced by Pitt and Warmuth (1990). Prediction-preserving reducibilities are a method of showing that one language is no harder to predict than another. Formally, let Lang$_1$ be a language over domain $X_1$ and Lang$_2$ be a language over domain $X_2$. We say that *predicting* Lang$_1$ *reduces to predicting* Lang$_2$, denoted Lang$_1 \trianglelefteq$ Lang$_2$, if there is a function $f_i : X_1 \to X_2$, henceforth called the *instance mapping*, and a function $f_c : $ Lang$_1 \to $ Lang$_2$, henceforth called the *concept mapping*, so that the following all hold:

1. $x \in C$ if and only if $f_i(x) \in f_c(C)$ — *i.e.*, concept membership is preserved by the mappings;

2. the size complexity of $f_c(C)$ is polynomial in the size complexity of $C$—*i.e.*, the size of concept representations is preserved within a polynomial factor;

3. $f_i(x)$ can be computed in polynomial time.

Note that $f_c$ need not be computable; also, since $f_i$ can be computed in polynomial time, $f_i(x)$ must also preserve size within a polynomial factor.

Intuitively, $f_c(C_1)$ returns a concept $C_2 \in$ Lang$_2$ that will "emulate" $C_1$—*i.e.*, make the same decisions about concept membership—on examples that have been "preprocessed" with the function $f_i$. If predicting Lang$_1$ reduces to predicting Lang$_2$ and a learning algorithm for Lang$_2$ exists, then one possible scheme for learning concepts from Lang$_1$ would be the following. First, convert any examples of the unknown concept $C_1$ from the domain $X_1$ to examples over the domain $X_2$ using the instance mapping $f_i$. If the conditions of the definition hold, then since $C_1$ is consistent with the original examples, the concept $f_c(C_1)$ will be consistent with their image under $f_i$; thus running the learning algorithm for Lang$_2$ should produce some hypothesis $H$ that is a good approximation of $f_c(C_1)$. Of course, it may not be possible to map $H$ back into the original language Lang$_1$, as computing $f_c{}^{-1}$ may be difficult or impossible. However, $H$ can still be used to predict membership in $C_1$: given an example $x$ from the original domain $X_1$, one can simply predict $x \in C_1$ to be true whenever $f_i(x) \in H$.

Pitt and Warmuth (1988) give a more rigorous argument that this approach leads to a prediction algorithm for Lang$_1$, leading to the following theorem.

**Theorem 1 (Pitt and Warmuth)** *Assume* LANG$_1$ $\unlhd$ LANG$_2$. *Then if* LANG$_1$ *is not polynomially predictable,* LANG$_2$ *is not polynomially predictable.*

## 3. Cryptographic Limitations on Learning Recursive Programs

Theorem 1 allows one to transfer hardness results from one language to another. This is useful because for a number of languages, it is known that prediction is as hard as breaking cryptographic schemes that are widely assumed to be secure. For example, it is known that predicting the class of languages accepted by deterministic finite state automata is "cryptographically hard", as is the class of languages accepted by log-space bounded Turing machines.

In this section we will make use of Theorem 1 and previous cryptographic hardness results to show that certain restricted classes of recursive logic programs are hard to learn.

### 3.1 Programs With $n$ Linear Recursive Clauses

In a companion paper (Cohen, 1995) we showed that a single linear closed recursive clause was identifiable from equivalence queries. In this section we will show that a program with a polynomial number of such clauses is not identifiable from equivalence queries, nor even polynomially predictable.

Specifically, let us extend our notion of a "family of languages" slightly, and let DLOG$[n,s]$ represent the language of log-space bounded deterministic Turing machines with up to $s$ states accepting inputs of size $n$ or less, with the usual semantics and complexity measure.[2] Also let $d$-DEPTH LIN REC PROG denote the family of logic programs containing only depth-$d$ linear closed recursive clauses, but containing any number of such clauses. We have the following result:

**Theorem 2** *For every $n$ and $s$, there exists a database $DB_{n,s} \in$ 1-$\mathcal{DB}$ and declaration $Dec_{n,s} \in$ 1-$\mathcal{Det}\mathcal{DEC}$ of sizes polynomial in $n$ and $s$ such that*

$$\mathrm{DLOG}[n,s] \ \unlhd \ 1\text{-DEPTH LIN REC PROG}[DB_{n,s}, Dec_{n,s}]$$

*Hence for $d \geq 1$ and $a \geq 1$, $d$-DEPTH LIN REC PROG$[\mathcal{DB}, a\text{-}\mathcal{Det}\mathcal{DEC}]$ is not uniformly polynomially predictable under cryptographic assumptions.*[3]

**Proof:** Recall that a log-space bounded Turing machine (TM) has an input tape of length $n$, a work tape of length $\log_2 n$ which initially contains all zeros, and a finite state control with state set $Q$. To simplify the proof, we assume without loss of generality that the tape and input alphabets are binary, that there is a single accepting state $q_f \in Q$, and that the machine will always erase its work tape and position the work tape head at the far left after it decides to accept its input.

At each time step, the machine will read the tape squares under its input tape head and work tape head, and based on these values and its current state $q$, it will

---

2. *I.e.*, a machine represents the set of all inputs that it accepts, and its complexity is the number of states.

3. Specifically, this language is not uniformly polynomially predictable unless all of the following cryptographic problems can be solved in polynomial time: solving the quadratic residue problem, inverting the RSA encryption function, and factoring Blum integers. This result holds because all of these cryptographic problems can be reduced to learning DLOG Turing machines (Kearns & Valiant, 1989).

- write either a 1 or a 0 on the work tape,

- shift the input tape head left or right,

- shift the work tape head left or right, and

- transition to a new internal state $q'$

A deterministic machine can thus be specified by a transition function

$$\delta : \{0,1\} \times \{0,1\} \times Q \longrightarrow \{0,1\} \times \{L, R\} \times \{L, R\} \times Q$$

Let us define the *internal configuration* of a TM to consist of the string of symbols written on the worktape, the position of the tape heads, and the internal state $q$ of the machine: thus a configuration is an element of the set

$$CON \equiv \{0,1\}^{\log_2 n} \times \{1, \ldots, \log_2 n\} \times \{1, \ldots, n\} \times Q$$

A simplified specification for the machine is the transition function

$$\delta' : \{0,1\} \times CON \rightarrow CON$$

where the component $\{0,1\}$ represents the contents of the input tape at the square below the input tape head.

Notice that for a machine whose worktape size is bounded by $\log n$, the cardinality of $CON$ is only $p = |Q|n^2 \log_2 n$, a polynomial in $n$ and $s = |Q|$. We will use this fact in our constructions.

The background database $DB_{n,s}$ is as follows. First, for $i = 0, \ldots, p$, an atom of the form $con_i(c_i)$ is present. Each constant $c_i$ will represent a different internal configuration of the Turing machine. We will also arbitrarily select $c_1$ to represent the (unique) accepting configuration, and add to $DB_{n,s}$ the atom $accepting(c_1)$. Thus

$$DB_{n,s} \equiv \{con_i(c_i)\}_{i=1}^{p} \cup \{accepting(c_1)\}$$

Next, we define the instance mapping. An instance in the Turing machine's domain is a binary string $X = b_1 \ldots b_n$; this is mapped by $f_i$ to the extended instance $(f, D)$ where

$$
\begin{aligned}
f &\equiv accepting(c_0) \\
D &\equiv \{true_i\}_{b_i \in X : b_i = 1} \quad \cup \quad \{false_i\}_{b_i \in X : b_i = 0}
\end{aligned}
$$

The description atoms have the effect of defining the predicate $true_i$ to be true iff the $i$-th bit of $X$ is a "1", and the defining the predicate $false_i$ to be true iff the $i$-th bit of $X$ is "0". The constant $c_0$ will represent the start configuration of the Turing machine, and the predicate $accepting(C)$ will be defined so that it is true iff the Turing machine accepts input $X$ starting from state $C$.

We will let $Dec_{n,s} = (accepting, 1, R)$ where $R$ contains the modes $con_i(+)$ and $con_i(-)$, for $i = 1, \ldots, p$; and $true_j$ and $false_j$ for $j = 1, \ldots, n$.

Finally, for the concept mapping $f_c$, let us assume some arbitrary one-to-one mapping $\eta$ between the internal configurations of a Turing machine $M$ and the predicate names $con_0, \ldots, con_{p-1}$ such that the start configuration $(0^{\log_2 n}, 1, q_0)$ maps to $con_0$ and the accepting configuration $(0^{\log_2 n}, 1, q_f)$ maps to $con_1$. We will construct the program $f_c(M)$ as follows. For each transition $\delta'(1, c) \to c'$ in $\delta'$, where $c$ and $c'$ are in $CON$, construct a clause of the form

$$\text{accepting}(C) \leftarrow \text{con}_j(C) \land \text{true}_i \land \text{con}_{j'}(C1) \land \text{accepting}(C1).$$

where $i$ is the position of the input tape head which is encoded in $c$, $con_j = \eta(c)$, and $con_{j'} = \eta(c')$. For each transition $\delta'(0, c) \to (c')$ in $\delta'$ construct an analogous clause, in which $true_i$ is replaced with $false_i$.

Now, we claim that for this program $P$, the machine $M$ will accept when started in configuration $c_i$ iff

$$DB_{n,s} \land D \land P \vdash accepting(c_i)$$

and hence that this construction preserves concept membership. This is perhaps easiest to see by considering the action of a top-down theorem prover when given the goal $accepting(C)$: the sequence of subgoals $accepting(c_i)$, $accepting(c_{i+1})$, ... generated by the theorem-prover precisely parallel the sequence of configurations $c_i$, ... entered by the Turing machine.

It is easily verified that the size of this program is polynomial in $n$ and $s$, and that the clauses are linear recursive, determinate, and of depth one, completing the proof. ∎

There are number of ways in which this result can be strengthened. Precisely the same construction used above can be used to reduce the class of nondeterministic log-space bounded Turing machines to the constant-depth determinate linear recursive programs. Further, a slight modification to the construction can be used to reduce the class of log-space bounded alternating Turing machines (Chandra, Kozen, & Stockmeyer, 1981) to constant-depth determinate 2-ary recursive programs. The modification is to emulate configurations corresponding to universal states of the Turing machine with clauses of the form

accepting(C) ←
    $\text{con}_j(C) \land \text{true}_i \land$
    $\text{con}_{j1'}(C1) \land \text{accepting}(C1) \land$
    $\text{con}_{j2'}(C2) \land \text{accepting}(C2).$

where $con_{j1'}$ and $con_{j2'}$ are the two successors to the universal configuration $con_j$. This is a very strong result, since log-space bounded alternating Turing machines are known to be able to perform every polynomial-time computation.

## 3.2 Programs With One $n$-ary Recursive Clause

We will now consider learning a single recursive clause with arbitrary closed recursion. Again, the key result of this section is an observation about expressive power: there is a background database that allows every log-space deterministic Turing machine $M$ to be emulated by a single recursive constant-depth determinate clause. This leads to the following negative predictability result.

**Theorem 3** *For every $n$ and $s$, there exists a database $DB_{n,s} \in 3\text{-}\mathcal{DB}$ and declaration $Dec_{n,s} \in 3\text{-}\mathcal{D}et\mathcal{DEC}$ of sizes polynomial in $n$ and $s$ such that*

$$\text{DLOG}[n,s] \trianglelefteq 3\text{-}\text{DEPTHREC}[DB_{n,s}, Dec_{n,s}]$$

*Hence for $d \geq 3$ and $a \geq 3$, $d\text{-}\text{DEPTHREC}[DB_n, a\text{-}\mathcal{D}et\mathcal{DEC}]$ is not uniformly polynomially predictable under cryptographic assumptions.*

**Proof:** Consider a DLOG machine $M$. As in the proof of Theorem 2, we assume without loss of generality that the tape alphabet is $\{0,1\}$, that there is a unique starting configuration $c_0$, and that there is a unique accepting configuration $c_1$. We will also assume without loss of generality that there is a unique "failing" configuration $c_{fail}$; and that there is exactly one transition of the form

$$\delta'(b, c_j) \rightarrow c'_j$$

for every combination of $i \in \{1, \ldots, n\}$, $b \in \{0,1\}$, and $c_j \in CON - \{c_1, c_{fail}\}$. Thus on input $X = x_1 \ldots x_n$ the machine $M$ starts with CONFIG=$c_0$, then executes transitions until it reaches CONFIG=$c_1$ or CONFIG=$c_{fail}$, at which point $X$ is accepted or rejected (respectively). We will use $p$ for the number of configurations. (Recall that $p$ is polynomial in $n$ and $s$.)

To emulate $M$, we will convert an example $X = b_1 \ldots b_n$ into the extended instance $f_i(X) = (f, D)$ where

$$\begin{aligned} f &\equiv accepting(c_0) \\ D &\equiv \{bit_i(b_i)\}_{i=1}^n \end{aligned}$$

Thus the predicate $bit_i(X)$ binds $X$ to the $i$-th bit of the TM's input tape. We also will define the following predicates in the background database $DB_{n,s}$.

- For every possible $b \in \{0,1\}$ and $j : 1 \leq j \leq p(n)$, the predicate $status_{b,j}(B,C,Y)$ will be defined so that given bindings for variables $B$ and $C$, $status_{b,j}(B,C,Y)$ will fail if $C = c_{fail}$; otherwise it will succeed, binding $Y$ to *active* if $B = b$ and $C = c_j$ and binding $Y$ to *inactive* otherwise.

- For $j : 1 \leq j \leq p(n)$, the predicate $next_j(Y,C)$ will succeed iff $Y$ can be bound to either *active* or *inactive*. If $Y = \oplus$, then $C$ will be bound to $c_j$; otherwise, $C$ will be bound to the accepting configuration $c_1$.

- The database also contains the fact $accepting(c_1)$.

It is easy to show that the size of this database is polynomial in $n$ and $s$.

The declaration $Dec_{n,s}$ is defined to be $(accepting, 1, R)$ where $R$ includes the modes $status_{bj}(+,+,-)$, $next_j(+,-)$, and $bit_i(-)$ for $b \in \{0,1\}$, $j = 1, \ldots, p$, and $i = 1, \ldots, n$.

Now, consider the transition rule $\delta'(b, c_j) \rightarrow c'_j$, and the corresponding conjunction

$$\text{TRANS}_{ibj} \equiv bit_i(B_{ibj}) \wedge status_{b,j}(C, B_{ibj}, Y_{ibj}) \wedge next_{j'}(Y_{ibj}, C1_{ibj}) \wedge accepting(C1_{ibj})$$

Given $DB_{n,s}$ and $D$, and assuming that $C$ is bound to some configuration $c$, this conjunction will fail if $c = c_{fail}$. It will succeed if $x_i \neq b$ or $c \neq c_j$; in this case $Y_{ibj}$ will be bound to *inactive*, $CI_{ibj}$ will be bound to $c_1$, and the recursive call succeeds because *accepting($c_1$)* is in $DB_{n,s}$. Finally, if $x_i = b$ and $c = c_j$, $\text{TRANS}_{ibj}$ will succeed only if the atom *accepting($c_{j'}$)* is provable; in this case, $Y_{ibj}$ will be bound to *active* and $CI_{ibj}$ will be bound to $c_{j'}$.

From this it is clear that the clause $f_c(M)$ below

$$\text{accepting(C)} \leftarrow \bigwedge_{\substack{i \in \{1,\dots,n\},\ b \in \{0,1\} \\ j \in \{1,\dots,p\}}} \text{TRANS}_{ibj}$$

will correctly emulate the machine $M$ on examples that have been preprocessed with the function $f_i$ described above. Hence this construction preserves concept membership. It is also easily verified that the size of this program is polynomial in $n$ and $s$, and that the clause is determinate and of depth three. ∎

## 3.3 One $k$-Local Linear Closed Recursive Clause

So far we have considered only one class of extensions to the positive result given in the companion paper (Cohen, 1995)—namely, relaxing the restrictions imposed on the recursive structure of the target program. Another reasonable question to ask is if linear closed recursive programs can be learned without the restriction of constant-depth determinacy.

In earlier papers (Cohen, 1993a, 1994a, 1993b) we have studied the conditions under which the constant-depth determinacy restriction can be relaxed while still allowing learnability for nonrecursive clauses. It turns out that most generalizations of constant-depth determinate clauses are not predictable, even without recursion. However, the language of nonrecursive clauses of constant *locality* is a pac-learnable generalization of constant-depth determinate clauses. Below, we will define this language, summarize the relevant previous results, and then address the question of the learnability of recursive local clauses.

Define a variable $V$ appearing in a clause $C$ to be *free* if it appears in the body of $C$ but not the head of $C$. Let $V_1$ and $V_2$ be two free variables appearing in a clause. $V_1$ *touches* $V_2$ if they appear in the same literal, and $V_1$ *influences* $V_2$ if it either touches $V_2$, or if it touches some variable $V_3$ that influences $V_2$. The *locale* of a free variable $V$ is the set of literals that either contain $V$, or that contain some free variable influenced by $V$. Informally, variable $V_1$ influences variable $V_2$ if the choice of a binding for $V_1$ can affect the possible choices of bindings for $V_2$.

The *locality* of a clause is the size of its largest locale. Let $k$-LOCALNONREC denote the language of nonrecursive clauses with locality $k$ or less. (That is, $k$-LOCALNONREC is the set of logic programs containing a single nonrecursive $k$-local clause.) The following facts are known (Cohen, 1993b):

- For fixed $k$ and $a$, the language family $k$-LOCALNONREC$[a\text{-}\mathcal{DB}, a\text{-}\mathcal{DEC}]$ is uniformly pac-learnable.

- For every constant $d$, every constant $a$, every database $DB \in a\text{-}\mathcal{DB}$, every declaration $Dec \in a\text{-}\mathcal{D}et\mathcal{DEC}$, and every clause $C \in d\text{-}\text{DEPTHNONREC}[DB, Dec]$, there is an equivalent clause $C'$ in $k$-LocalNonRec$[DB, Dec]$ of size bounded by $k\|C\|$, where $k$ is a function only of $a$ and $d$ (and hence is a constant if $d$ and $a$ are also constants.)

Hence

$$k\text{-}\textsc{LocalNonRec}[\mathcal{DB}, a\text{-}\mathcal{DEC}]$$

is a pac-learnable generalization of

$$d\text{-}\textsc{DepthNonRec}[\mathcal{DB}, a\text{-}\mathcal{DetDEC}]$$

It is thus plausible to ask if recursive programs of $k$-local clauses are pac-learnable. Some facts about the learnability of $k$-local programs follow immediately from previous results. For example, an immediate consequence of the construction of Theorem 2 is that programs with a polynomial number of linear recursive $k$-local clauses are not predictable for $k \geq 2$. Similarly, Theorem 3 shows that a single recursive $k$-local clause is not predictable for $k \geq 4$.

It is still reasonable to ask, however, if the positive result for bounded-depth determinate recursive clauses (Cohen, 1995) can be extended to $k$-ary closed recursive $k$-local clauses. Unfortunately, we have the following negative result, which shows that even linear closed recursive clauses are not learnable.

**Theorem 4** *Let* Dfa$[s]$ *denote the language of deterministic finite automata with $s$ states, and let $k$-LocalLinRec be the set of linear closed recursive $k$-local clauses. For any constant $s$ there exists a database $DB_s \in 3\text{-}\mathcal{DB}$ and a declaration $Dec_s \in 3\text{-}\mathcal{DEC}$, both of size polynomial in $s$, such that*

$$\textsc{Dfa}[s] \trianglelefteq 3\text{-}\textsc{LocalLinRec}[DB_s, Dec_s]$$

*Hence for $k \geq 3$ and $a \geq 3$, $k$-LocalLinRec$[a\text{-}\mathcal{DB}, Dec]$ is not uniformly polynomially predictable under cryptographic assumptions.*

**Proof:** Following Hopcroft and Ullman (1979) we will represent a DFA $M$ over the alphabet $\Sigma$ as a tuple $(q_0, Q, F, \delta)$ where $q_0$ is the initial state, $Q$ is the set of states, $F$ is the set of accepting states, and $\delta : Q \times \Sigma \to Q$ is the transition function (which we will sometimes think of as a subset of $Q \times \Sigma \times Q$). To prove the theorem, we need to construct a database $DB_s$ of size polynomial in $s$ such that every $s$-state DFA can be emulated by a linear recursive $k$-local clause over $DB_s$.

Rather than directly emulating $M$, it will be convenient to emulate instead a modification of $M$. Let $\hat{M}$ be a DFA with state set $\hat{Q} \equiv Q \cup \{q_{(-1)}, q_e, q_f\}$, where $q_{(-1)}$, $q_e$ and $q_f$ are new states not found in $Q$. The initial state of $\hat{M}$ is $q_{(-1)}$. The only final state of $\hat{M}$ is $q_f$. The transition function of $\hat{M}$ is

$$\hat{\delta} \equiv \delta \cup \{(q_{(-1)}, a, q_0), (q_e, c, q_f)\} \cup \bigcup_{q_i \in F} \{(q_i, b, q_e)\}$$

where $a$, $b$, and $c$ are new letters not in $\Sigma$. Note that $\hat{M}$ is now a DFA over the alphabet $\Sigma \cup \{a, b, c\}$, and, as described, need not be a complete DFA over this alphabet. (That is, there may be pairs $(q_i, a)$ such that $\hat{\delta}(q_i, a)$ is undefined.) However, $\hat{M}$ can be easily made complete by introducing an additional rejecting state $q_r$, and making every undefined transition lead to $q_r$. More precisely, let $\delta'$ be defined as

$$\delta' \equiv \hat{\delta} \cup \{(q_i, x, q_r) \mid q_i \in \hat{Q} \;\wedge\; x \in \Sigma \cup \{a, b, c\} \;\wedge\; (\not\exists q_j : (q_i, x, q_j) \in \hat{\delta})\}$$

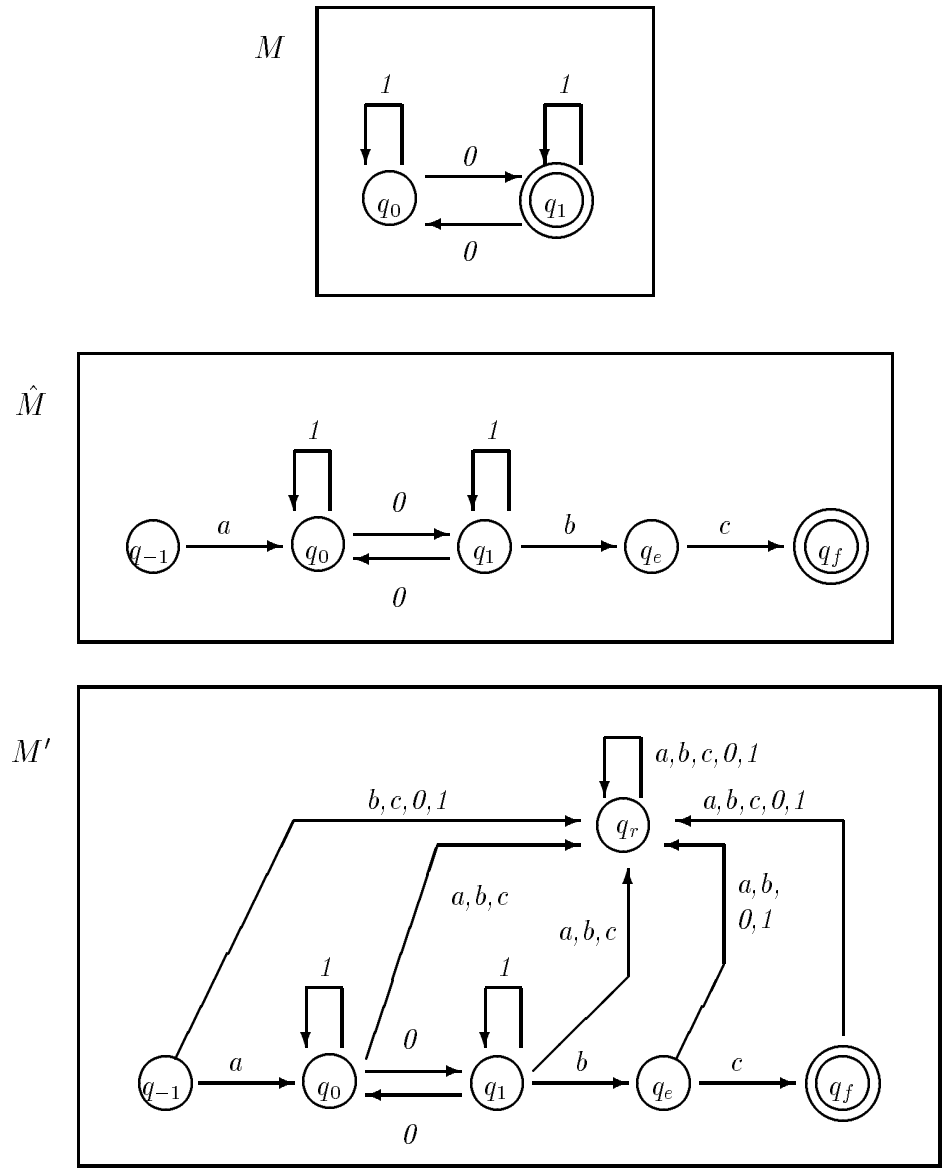

Figure 1: How a DFA is modified before emulation with a local clause

Thus $M' = (q_{(-1)}, \hat{Q} \cup \{q_r\}, \{q_f\}, \delta')$ is a "completed" version of $\hat{M}$, with $Q' = \hat{Q} \cup \{q_r\}$. We will use $M'$ in the construction below; we will also let $Q' = \hat{Q} \cup \{q_r\}$ and $\Sigma' = \Sigma \cup \{a, b, c\}$.

Examples of $M$, $\hat{M}$ and $M'$ are shown in Figure 1. Notice that aside from the arcs into and out of the rejecting state $q_r$, the state diagram of $M'$ is nearly identical to that of $M$. The differences are that in $M'$ there is a new initial state $q_{(-1)}$ with a single outgoing arc labeled $a$ to the old initial state $q_0$; also every final state of $M$ has in $M'$ an outgoing arc labeled $b$ to a new state $q_e$, which in turn has a single outgoing arc labeled $c$ to the final state $q_f$. It is easy to show that

$$x \in L(M) \text{ iff } axbc \in L(M')$$

Now, given a set of states $Q'$ we define a database $DB$ that contains the following predicates:

- $arc_{q_i, \sigma, q_j}(S, X, T)$ is true for any $S \in Q'$, any $T \in Q'$, and any $X \in \Sigma'$, unless $S = q_i$, $X = \sigma$, and $T \neq q_j$.

- $state(S)$ is true for any $S \in Q'$.

- $accept(c, nil, q_e, q_f)$ is true.

As motivation for the $arc$ predicates, observe that in emulating $M'$ it is clearly useful to be able to represent the transition function $\delta'$. The usefulness of the $arc$ predicates is that any transition function $\delta'$ can be represented using a conjunction of $arc$ literals. In particular, the conjunction

$$\bigwedge_{(q_i, \sigma, q_j) \in \delta'} arc_{q_i, \sigma, q_j}(S, X, T)$$

succeeds when $\delta'(S, X) = T$, and fails otherwise.

Let us now define the instance mapping $f_i$ as $f_i(x) = (f, D)$ where

$$f = accept(a, xbc, q_{(-1)}, q_0)$$

and $D$ is a set of facts that defines the *components* relation on the list that corresponds to the string $xbc$. In other words, if $x = \sigma_1 \ldots \sigma_n$, then $D$ is the set of facts

components($\sigma_1 \ldots \sigma_n bc, \sigma_1, \sigma_2 \ldots \sigma_n bc$)
components($\sigma_2 \ldots \sigma_n bc, \sigma_2, \sigma_3 \ldots \sigma_n bc$)
$\vdots$
components(c,c,nil)

The declaration $Dec_n$ will be $Dec_n = (accept, 4, R)$ where $R$ contains the modes $components(+, -, -)$, $state(-)$, and $arc_{q_i, \sigma, q_j}(+, +, +)$ for $q_i$, $q_j$ in $Q'$, and $\sigma \in \Sigma'$.

Finally, define the concept mapping $f_c(M)$ for a machine $M$ to be the clause

accept(X,Ys,S,T) ←
   $\bigwedge_{(q_i,\sigma,q_j)\in\delta'}$ arc$_{q_i,\sigma,q_j}$(S,X,T)
   ∧ components(Ys,X1,Ys1) ∧ state(U) ∧ accept(X1,Ys1,T,U).

where $\delta'$ is the transition function for the corresponding machine $M'$ defined above. It is easy to show this construction is polynomial.

In the clause $X$ is a letter in $\Sigma'$, $Ys$ is a list of such letters, and $S$ and $T$ are both states in $Q'$. The intent of the construction is that the predicate *accept* will succeed exactly when (a) the string $XYs$ is accepted by $M'$ when $M'$ is started in state $S$, and (b) the first action taken by $M'$ on the string $XYs$ is to go from state $S$ to state $T$.

Since all of the initial transitions in $M'$ are from $q_{(-1)}$ to $q_0$ on input $a$, then if the predicate *accept* has the claimed behavior, clearly the proposed mapping satisfies the requirements of Theorem 1. To complete the proof, therefore, we must now verify that the predicate *accept* succeeds iff $XYs$ is accepted by $M'$ in state $S$ with an initial transition to $T$.

From the definition of DFAs the string $XYs$ is accepted by $M'$ in state $S$ with an initial transition to $T$ iff one of the following two conditions holds.

- $\delta'(S,X) = T$, $Ys$ is the empty string and $T$ is a final state of $M'$, or;

- $\delta'(S,X) = T$, $Ys$ is a nonempty string (and hence has some head $X1$ and some tail $Ys1$) and $Ys1$ is accepted by $M'$ in state $T$, with any initial transition.

The base fact *accept(c,nil,q_e,q_f)* succeeds precisely when the first case holds, since in $M'$ this transition is the only one to a final state. In the second case, the conjunction of the *arc* conditions in the $f_c(M)$ clause succeeds exactly when $\delta(S,X) = T$ (as noted above). Further the second conjunction in the clause can be succeeds when $Ys$ is a nonempty string with head $X1$ and tail $Ys1$ and $X1Ys1$ is accepted by $M'$ in state $T$ with initial transition to any state $U$, which corresponds exactly to the second case above.

Thus concept membership is preserved by the mapping. This completes the proof. ■

## 4. DNF-Hardness Results for Recursive Programs

To summarize previous results for determinate clauses, it was shown that while a single $k$-ary closed recursive depth-$d$ clause is pac-learnable (Cohen, 1995), a set of $n$ linear closed recursive depth-$d$ clauses is not; further, even a single $n$-ary closed recursive depth-$d$ clauses is not pac-learnable. There is still a large gap between the positive and negative results, however: in particular, the learnability of recursive programs containing a constant number of $k$-ary recursive clauses has not yet been established.

In this section we will investigate the learnability of these classes of programs. We will show that programs with either two linear closed recursive clauses or one linear closed recursive clause and one base case are as hard to learn as boolean functions in disjunctive normal form (DNF). The pac-learnability of DNF is a long-standing open problem in computational learning theory; the import of these results, therefore, is that establishing the learnability of these classes will require some substantial advance in computational learning theory.

## 4.1 A Linear Recursive Clause Plus a Base Clause

Previous work has established that two-clause constant-depth determinate programs consisting of one linear recursive clause and one nonrecursive clause can be identified, given two types of oracles: the standard equivalence-query oracle, and a "basecase oracle' (Cohen, 1995). (The basecase oracle determines if an example is covered by the nonrecursive clause alone.) In this section we will show that in the absence of the basecase oracle, the learning problem is as hard as learning boolean DNF.

In the discussion below, $\text{DNF}[n, r]$ denotes the language of $r$-term boolean functions in disjunctive normal form over $n$ variables.

**Theorem 5** *Let $d$-DEPTH-2-CLAUSE be the set of 2-clause programs consisting of one clause in $d$-DEPTHLINREC and one clause in $d$-DEPTHNONREC. Then for any $n$ and any $r$ there exists a database $DB_{n,r} \in 2\text{-}\mathcal{DB}$ and a declaration $Dec_{n,r} \in 2\text{-}\mathcal{DEC}$, both of sizes polynomial in $n$ and $r$, such that*

$$\text{DNF}[n, r] \trianglelefteq 1\text{-DEPTH-2-CLAUSE}[DB_{n,r}, Dec_{n,r}]$$

*Hence for $a \geq 2$ and $d \geq 1$ the language family $d$-DEPTH-2-CLAUSE$[\mathcal{DB}, a\text{-}\mathcal{Det}\mathcal{DEC}]$ is uniformly polynomially predictable only if DNF is polynomially predictable.*

**Proof:** We will produce a $DB_{n,r} \in \mathcal{DB}$ and $Dec_{n,r} \in 2\text{-}\mathcal{Det}\mathcal{DEC}$ such that predicting DNF can be reduced to predicting 1-DEPTH-2-CLAUSE$[DB_{n,r}, Dec_{n,r}]$. The construction makes use of a trick first used in Theorem 3 of (Cohen, 1993a), in which a DNF formula is emulated by a conjunction containing a single variable $Y$ which is existentially quantified over a restricted range.

We begin with the instance mapping $f_i$. An assignment $\eta = b_1 \ldots b_n$ will be converted to the extended instance $(f, D)$ where

$$
\begin{aligned}
f &\equiv p(1) \\
D &\equiv \{bit_i(b_i)\}_{i=1}^n
\end{aligned}
$$

Next, we define the database $DB_{n,r}$ to contain the binary predicates $true_1, false_1, \ldots, true_r$, $false_r$ which behave as follows:

- $true_i(X, Y)$ succeeds if $X = 1$, or if $Y \in \{1, \ldots, r\} - \{i\}$.

- $false_i(X, Y)$ succeeds if $X = 0$, or if $Y \in \{1, \ldots, r\} - \{i\}$.

Further, $DB_{n,r}$ contains facts that define the predicate $succ(Y, Z)$ to be true whenever $Z = Y + 1$, and both $Y$ and $Z$ are numbers between 1 and $r$. Clearly the size of $DB_{n,r}$ is polynomial in $r$.

Let $Dec_{n,r} = (p, 1, R)$ where $R$ contains the modes $bit_i(-)$, for $i = 1, \ldots, n$; $true_j(+, +)$ and $false_j(+, +)$, for $j = 1, \ldots, r$, and $succ(+, -)$.

Now let $\phi$ be an $r$-term DNF formula $\phi = \vee_{i=1}^r \wedge_{j=1}^{s_i} l_{ij}$ over the variables $v_1, \ldots, v_n$. We may assume without loss of generality that $\phi$ contains exactly $r$ terms, since any DNF formula with fewer than $r$ terms can be padded to exactly $r$ terms by adding terms of the

**Background database:**

  for $i = 1, \ldots, r$

   $true_i(b, y)$  for all $b, y : b = 1$ or $y \in \{1, \ldots, r\}$ but $y \neq i$

   $false_i(b, y)$  for all $b, y : b = 0$ or $y \in \{1, \ldots, r\}$ but $y \neq i$

  succ(y,z)    if $z = y + 1$ and $y \in \{1, \ldots, r\}$ and $z \in \{1, \ldots, r\}$

**DNF formula:**   $(v_1 \wedge \overline{v_3} \wedge v_4) \ \vee \ (\overline{v_2} \wedge \overline{v_3}) \ \vee \ (v_1 \wedge \overline{v_4})$

**Equivalent program:**

  p(Y) ←succ(Y,Z)∧p(Z).

  p(Y) ←bit₁(X₁) ∧ bit₂(X₂) ∧ bit₃(X₃) ∧ bit₄(X₄) ∧

   true₁(X₁,Y) ∧ false₁(X₃,Y) ∧ true₁(X₄,Y) ∧

   false₂(X₂,Y) ∧ false₂(X₃,Y)∧

   true₃(X₁,Y) ∧ false₃(X₄,Y).

**Instance mapping:**   $f_i(1011) = (p(1), \{bit_1(1), bit_2(0), bit_3(1), bit_4(1)\})$

Figure 2: Reducing DNF to a recursive program

form $v_1 \overline{v_1}$. We now define the concept mapping $f_c(\phi)$ to be the program $C_R, C_B$ where $C_R$ is the linear recursive depth 1 determinate clause

$$p(Y) \leftarrow succ(Y, Z) \wedge p(Z)$$

and $C_B$ is the nonrecursive depth 1 determinate clause

$$p(Y) \leftarrow \bigwedge_{k=1}^{n} bit_k(X_k) \ \wedge \ \bigwedge_{i=1}^{r} \bigwedge_{j=1}^{s_i} B_{ij}$$

where $B_{ij}$ is defined as follows:

$$B_{ij} \equiv \begin{cases} true_i(X_k, Y) & \text{if } l_{ij} = v_k \\ false_i(X_k, Y) & \text{if } l_{ij} = \overline{v_k} \end{cases}$$

An example of the construction is shown in Figure 2; we suggest that the reader refer to this figure at this point. The basic idea behind the construction is that first, the clause $C_B$ will succeed only if the variable $Y$ is bound to $i$ and the $i$-th term of $\phi$ succeeds (the definitions of $true_i$ and $false_i$ are designed to ensure that this property holds); second, the recursive clause $C_R$ is constructed so that the program $f_c(\phi)$ succeeds iff $C_B$ succeeds with $Y$ bound to one of the values $1, \ldots, n$.

We will now argue more rigorously for the correctness of the construction. Clearly, $f_i(\eta)$ and $f_c(\phi)$ are of the same size as $\eta$ and $\phi$ respectively. Since $DB_{n,r}$ is also of polynomial size, this reduction is polynomial.

Figure 3 shows the possible proofs that can be constructed with the program $f_c(\phi)$; notice that the program $f_c(\phi)$ succeeds exactly when the clause $C_B$ succeeds for some value

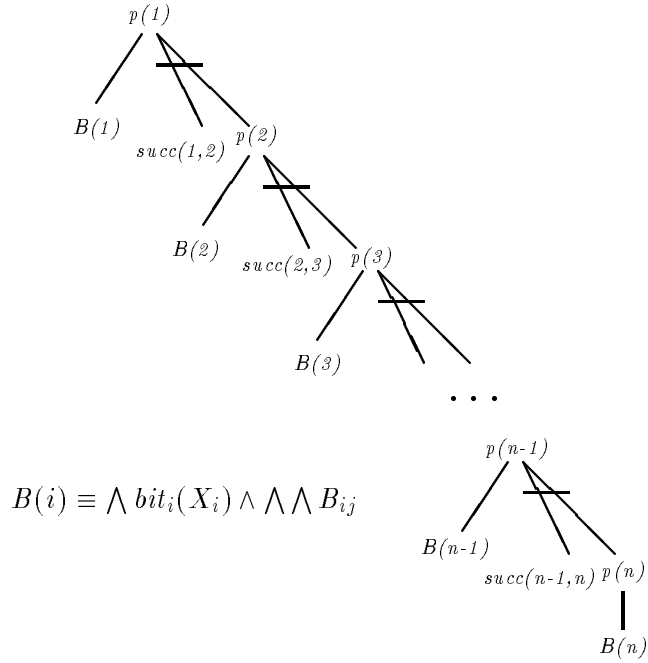

$$B(i) \equiv \bigwedge bit_i(X_i) \wedge \bigwedge \bigwedge B_{ij}$$

Figure 3: Space of proofs possible with the program $f_c(\phi)$

of $Y$ between 1 and $r$. Now, if $\phi$ is true then some term $T_i = \bigwedge_{j=1}^{s_i} l_{ij}$ must be true; in this case $\bigwedge_{j=1}^{s_i} B_{ij}$ succeeds with $Y$ bound to the value $i$ and $\bigwedge_{j=1}^{s_{i'}} B_{i'j}$ for every $i' \neq i$ also succeeds with $Y$ bound to $i$. On the other hand, if $\phi$ is false for an assignment, then each $T_i$ fails, and hence for every possible binding of $Y$ generated by repeated use of the recursive clause $C_R$ the base clause $C_B$ will also fail. Thus concept membership is preserved by the mapping.

This concludes the proof. ∎

## 4.2 Two Linear Recursive Clauses

Recall again that a single linear closed recursive clause is identifiable from equivalence queries (Cohen, 1995). A construction similar to that used in Theorem 5 can be used to show that this result cannot be extended to programs with two linear recursive clauses.

**Theorem 6** *Let $d$-DEPTH-2-CLAUSE' be the set of 2-clause programs consisting of two clauses in $d$-DEPTHLINREC. (Thus we assume that the base case of the recursion is given as background knowledge.) Then for any constants $n$ and $r$ there exists a database $DB_{n,r} \in$ 2-$\mathcal{DB}$ and a declaration $Dec_{n,r} \in$ 2-$\mathcal{DEC}$, both of sizes polynomial in $n$, such that*

$$\text{DNF}[n,r] \trianglelefteq 1\text{-DEPTH-2-CLAUSE}'[DB_{n,r}, Dec_{n,r}]$$

*Hence for any constants $a \geq 2$ and $d \geq 1$ the language family*

$$d\text{-DEPTH-2-CLAUSE}'[\mathcal{DB}, a\text{-}\mathcal{D}et\mathcal{DEC}]$$

*is uniformly polynomially predictable only if DNF is polynomially predictable.*

**Proof:** As before, the proof makes use of a prediction-preserving reducibility from DNF to $d$-Depth-2-Clause$'[DB, Dec]$ for a specific $DB$ and $Dec$. Let us assume that $\phi$ is a DNF with $r$ terms, and further assume that $r = 2^k$. (Again, this assumption is made without loss of generality, since the number of terms in $\phi$ can be increased by padding with vacuous terms.) Now consider a complete binary tree of depth $k + 1$. The $k$-th level of this tree has exactly $r$ nodes; let us label these nodes $1, \ldots, r$, and give the other nodes arbitrary labels. Now construct a database $DB_{n,r}$ as in Theorem 5, except for the following changes:

- The predicates $true_i(b, y)$ and $false_i(b, y)$ also succeed when $y$ is the label of a node at some level below $k$.

- Rather than the predicate $succ$, the database contains two predicates $leftson$ and $rightson$ that encode the relationship between nodes in the binary tree.

- The database includes the facts $p(\omega_1), \ldots, p(\omega_{2r})$, where $\omega_1, \ldots, \omega_{2r}$ are the leaves of the binary tree. These will be used as the base cases of the recursive program that is to be learned.

Let $\rho$ be the label of the root of the binary tree. We define the instance mapping to be

$$f_i(b_1 \ldots b_1) \equiv (p(\rho), \{bit_1(b_1), \ldots, bit_n(b_n)\})$$

Note that except for the use of $\rho$ rather than 1, this is identical to the instance mapping used in Theorem 5. Also let $Dec_{n,r} = (p, 1, R)$ where $R$ contains the modes $bit_i(-)$, for $i = 1, \ldots, n$; $true_j(+, +)$ and $false_j(+, +)$, for $j = 1, \ldots, r$; $leftson(+, -)$; and $rightson(+, -)$.

The concept mapping $f_c(\phi)$ is the pair of clauses $R_1, R_2$, where $R_1$ is the clause

$$p(Y) \leftarrow \bigwedge_{k=1}^{n} bit_k(X_k) \; \wedge \; \bigwedge_{i=1}^{r} \bigwedge_{j=1}^{s_i} B_{ij} \; \wedge \; leftson(Y, Z) \wedge p(Z)$$

and $R_2$ is the clause

$$p(Y) \leftarrow \bigwedge_{k=1}^{n} bit_k(X_k) \; \wedge \; \bigwedge_{i=1}^{r} \bigwedge_{j=1}^{s_i} B_{ij} \; \wedge \; rightson(Y, Z) \wedge p(Z)$$

Note that both of these clause are linear recursive, determinate, and have depth 1. Also, the construction is clearly polynomial. It remains to show that membership is preserved.

Figure 4 shows the space of proofs that can be constructed with the program $f_c(\phi)$; as in Figure 3, $B(i)$ abbreviates the conjunction $\bigwedge bit_i(X_i) \wedge \bigwedge \bigwedge B_{ij}$. Notice that the program will succeed only if the recursive calls manage to finally recurse to one of the base cases $p(\omega_1), \ldots, p(\omega_{2r})$, which correspond to the leaves of the binary tree. Both clauses will both succeed on the the first $k - 1$ levels of the tree. However, to reach the base cases of the recursion at the leaves of the tree, the recursion must pass through the $k$-th level of the tree; that is, one of the clauses above must succeed on some node $y$ of the binary tree, where $y$ is on the $k$-th level of the tree, and hence the label of $y$ is a number between 1 and $r$. The program thus succeeds on $f_i(\eta)$ precisely when there is some number $y$ between 1 and $r$ such that the conjunction $B(i)$ succeeds, which (by the argument given in Theorem 5) can happen if and only if $\phi$ is satisfied by the assignment $\eta$. Thus, the mappings preserve concept membership. This completes the proof. ∎

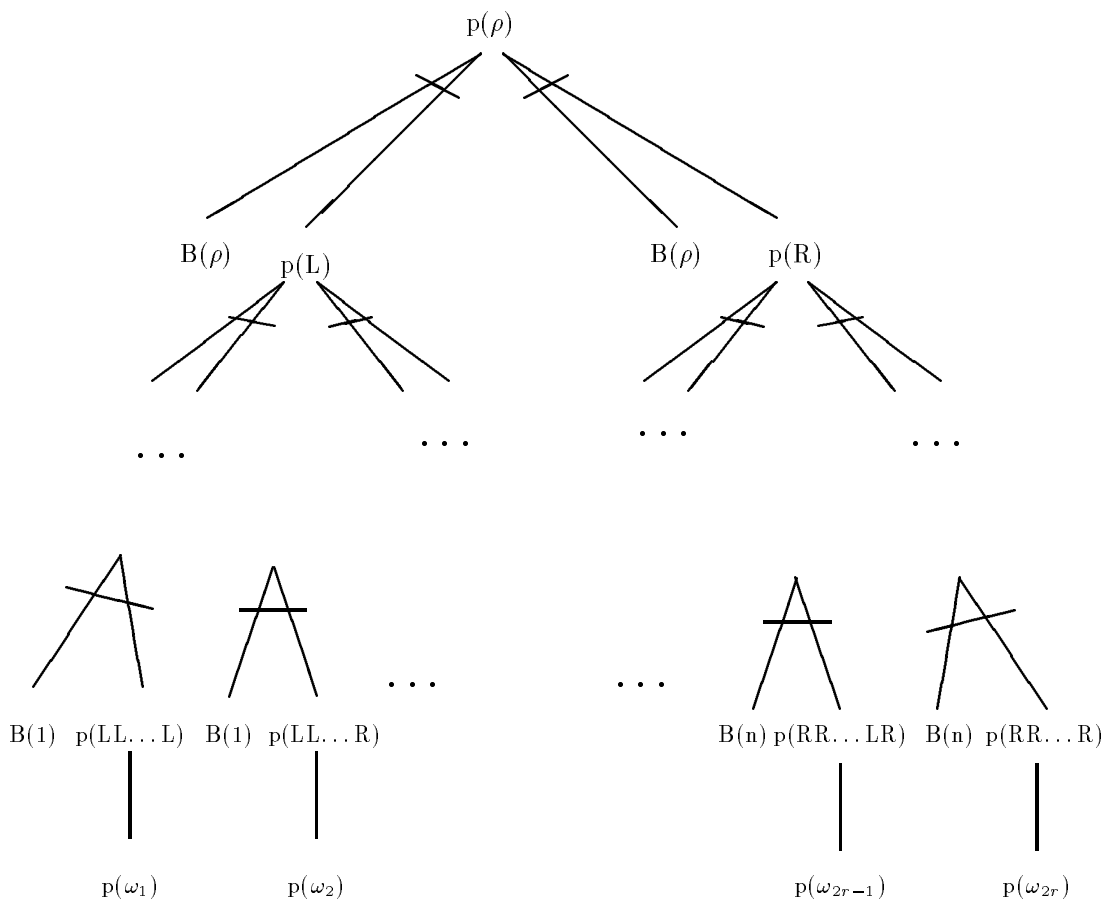

Figure 4: Proofs possible with the program $f_c(\phi)$

Notice that the programs $f_c(\phi)$ used in this proof all have the property that the depth of every proof is logarithmic in the size of the instances. This means that the hardness result holds even if one additionally restricts the class of programs to have a logarithmic depth bound.

## 4.3 Upper Bounds on the Difficulty of Learning

The previous sections showed that several highly restricted classes of recursive programs are at least as hard to predict as DNF. In this section we will show that these restricted classes are also no harder to predict than DNF.

We will wish to restrict the depth of a proof constructed by a target program. Thus, let $h(n)$ be any function; we will use $\text{LANG}_{h(n)}$ for the set of programs in the class $\text{LANG}$ such that all proofs of an extended instance $(f, D)$ have depth bounded by $h(\|D\|)$.

**Theorem 7** *Let* DNF$[n, *]$ *be the language of DNF boolean functions (with any number of terms), and recall that* $d$-DEPTH-$2$-CLAUSE *is the language of $2$-clause programs consisting of one clause in* $d$-DEPTHLINREC *and one clause in* $d$-DEPTHNONREC, *and that* $d$-DEPTH-$2$-CLAUSE$'$ *is the language of $2$-clause programs consisting of two clauses in* $d$-DEPTHLINREC.

*For all constants $d$ and $a$, and all databases $DB \in \mathcal{DB}$ and declarations $Dec \in a$-$\mathcal{DetDEC}$, there is a polynomial function* $poly(n)$ *such that*

- $d$-DEPTH-$2$-CLAUSE$[DB, Dec] \trianglelefteq$ DNF$[poly(\|DB\|), *]$

- $d$-DEPTH-$2$-CLAUSE$'_{h(n)}[DB, Dec] \trianglelefteq$ DNF$[poly(\|DB\|), *]$ *if $h(n)$ is bounded by $c \log n$ for some constant $c$.*

*Hence if either of these language families is uniformly polynomially predictable, then* DNF$[n, *]$ *is polynomially predictable.*

**Proof:** The proof relies on several facts established in the companion paper (Cohen, 1995).

- For every declaration $Dec$, there is a clause $BOTTOM^*_d(Dec)$ such that every nonrecursive depth-$d$ determinate clause $C$ is equivalent to some subclause of $BOTTOM^*_d$. Further, the size of $BOTTOM^*d$ is polynomial in $Dec$. This means that the language of subclauses of $BOTTOM^*$ is a normal form for nonrecursive constant-depth determinate clauses.

- Every linear closed recursive clause $C_R$ that is constant-depth determinate is equivalent to some subclause of $BOTTOM^*$ plus a recursive literal $L_r$; further, there are only a polynomial number of possible recursive literals $L_r$.

- For any constants $a$, $a'$, and $d$, any database $DB \in a$-$\mathcal{DB}$, any declaration $Dec = (p, a', R)$, any database $DB \in a$-$\mathcal{DB}$, and any program $P$ in $d$-DEPTH-$2$-CLAUSE$[DB, Dec]$, the depth of a terminating proof constructing using $P$ is no more than $h_{\max}$, where $h_{\max}$ is a polynomial in the size of $DB$ and $Dec$.

- At can be assumed without loss of generality that the database $DB$ and all decsriptions $D$ contain an *equality predicate*, where an equality predicate is simply a predicate $equal(X, Y)$ which is true exactly when $X = Y$.

The idea of the proof is to contruct a prediction-preserving reduction between the two classes of recursive programs listed above to and DNF. We will begin with two lemmas.

**Lemma 8** *Let $Dec \in a$-$\mathcal{DetDEC}$, and let $C$ be a nonrecursive depth-$d$ determinate clause consistent with $Dec$. Let* SUBCLAUSE$_C$ *denote the language of subclauses of $C$, and let* MONOMIAL$[u]$ *denote the language of monomials over $u$ variables. Then there is a polynomial $poly_1$ so that for any database $DB \in \mathcal{DB}$,*

$$\text{SUBCLAUSE}_C[DB, Dec] \trianglelefteq \text{MONOMIAL}[poly_1(\|DB\|)]$$

**Proof of lemma:** Follows immediately from the construction used in Theorem 1 of Džeroski, Muggleton, and Russell (Džeroski et al., 1992). (The basic idea of the construction is to introduce a propositional variable representing the "success" of each connected chain of literals in $C$. Any subclause of $C$ can then be represented as a conjunction of these propositions.) ∎

This lemma can be extended as follows.

**Lemma 9** *Let $Dec \in a\text{-}\mathcal{D}et\mathcal{DEC}$, and let $S = \{C_1, \ldots, C_r\}$ be a set of $r$ nonrecursive depth-$d$ determinate clauses consistent with $Dec$, each of length $n$ or less. Let $\textsc{Subclause}_S$ denote the set of all programs of the form $P = (D_1, \ldots, D_s)$ such that each $D_i$ is a subclause of some $C_j \in S$.*

*Then there is a polynomial $poly_2$ so that for any database $DB \in \mathcal{DB}$,*

$$\textsc{Subclause}_S[DB, Dec] \trianglelefteq \textsc{Dnf}[poly_2(\|DB\|, r), *]$$

**Proof of lemma:** By Lemma 8, for each $C_i \in S$, there is a set of variables $V_i$ of size polynomial in $\|DB\|$ such that every clause in $\textsc{Subclause}_{C_i}$ can be emulated by a monomial over $V_i$. Let $V = \bigcup_{i=1}^{r} V_i$. Clearly, $|V|$ is polynomial in $n$ and $r$, and every clause in $\bigcup_i \textsc{Subclause}_{C_i}$ can be also emulated by a monomial over $V$. Further, every disjunction of $r$ such clauses can be represented by a disjunction of such monomials.

Since the $C_i$'s all satisfy a single declaration $Dec = (p, a, R)$, they have heads with the same principle function and arity; further, we may assume (without loss of generality, since an equality predicate is assumed) that the variables appearing in the heads of these clauses are all distinct. Since the $C_i$'s are also nonrecursive, every program $P \in \textsc{Subclause}_S$ can be represented as a disjunction $D_1 \vee \ldots \vee D_r$ where for all $i$, $D_i \in (\bigcup_i \textsc{Subclause}_{C_i})$. Hence every $P \in \textsc{Subclause}_S$ can be represented by an $r$-term DNF over the set of variables $V$. ∎

Let us now introduce some additional notation. If $C$ and $D$ are clauses, then we will use $C \sqcap D$ to denote the result of resolving $C$ and $D$ together, and $C^i$ to denote the result of resolving $C$ with itself $i$ times. Note that $C \sqcap D$ is unique if $C$ is linear recursive and $C$ and $D$ have the same predicate in their heads (since there will be only one pair of complementary literals.)

Now, consider some target program

$$P = (C_R, C_B) \in d\text{-}\textsc{Depth-2-Clause}[DB, Dec]$$

where $C_R$ is the recursive clause and $C_B$ is the base. The proof of any extended instance $(f, D)$ must use clause $C_R$ repeatedly $h$ times and then use clause $C_B$ to resolve away the final subgoal. Hence the nonrecursive clause $C_R^h \sqcap C_B$ could also be used to cover the instance $(f, D)$.

Since the depth of any proof for this class of programs is bounded by a number $h_{max}$ that is polynomial in $\|DB\|$ and $n_e$, the nonrecursive program

$$P' = \{C_R^h \sqcap C_B : 0 \le h \le h_{max}\}$$

is equivalent to $P$ on extended instances of size $n_e$ or less.

Finally, recall that we can assume that $C_B$ is a subclause of $BOTTOM_d^*$; also, there is a polynomial-sized set $L_{\mathcal{R}} = L_{r_1}, \ldots, L_{r_p}$ of closed recursive literals such that for some $L_{r_i} \in L_{\mathcal{R}}$, the clause $C_R$ is a subclause of $BOTTOM_d^* \cup L_{r_i}$. This means that if we let $S$ be the polynomial-sized set

$$S_1 = \{(BOTTOM_d^* \cup L_{r_i})^h \sqcap BOTTOM_d^* \mid 0 \leq h \leq h_{max} \text{ and } L_{r_i} \in L_{\mathcal{R}}\}$$

then $P' \in \text{Subclause}_{S_1}$. Thus by Lemma 9, $d$-Depth-2-Clause $\trianglelefteq$ Dnf. This concludes the proof of the first statement in the the theorem.

To show that

$$d\text{-Depth-2-Clause}'_{h(n)}[DB, Dec] \trianglelefteq \text{Dnf}[poly(\|DB\|, *]$$

a similar argument applies. Let us again introduce some notation, and define $\text{MESH}_{h,n}(C_{R_1}, C_{R_2})$ as the set of all clauses of the form

$$C_{R_{i,1}} \sqcap C_{R_{i,2}} \sqcap \ldots \sqcap C_{R_{i,h'}}$$

where for all $j$, $C_{R_{ij}} = C_{R_1}$ or $C_{R_{ij}} = C_{R_2}$, and $h' \leq h(n)$. Notice that for functions $h(n) \leq c \log n$ the number of such clauses is polynomial in $n$.

Now let $p$ be the predicate appearing in the heads of $C_{R1}$ and $C_{R_2}$, and let $\hat{C}$ (respectively $\hat{DB}$) be a a version of $C$ ($DB$) in which every instance of the predicate $p$ has been replaced with a new predicate $\hat{p}$. If $P$ is a recursive program $P = \{C_{R_1}, C_{R_2}\}$ in $d$-Depth-2-Clause$'$ over the database $DB$, then $P \wedge DB$ is equivalent[4] to the nonrecursive program $P' \wedge \hat{DB}$, where

$$P' = \{\hat{C} \mid C \in \text{MESH}_{h,n_e}(C_{R_1}, C_{R_2})\}$$

Now recall that there are a polynomial number of recursive literals $L_{r_i}$, and hence a polynomial number of pairs of recursive literals $L_{r_i}, L_{r_j}$. This means that the set of clauses

$$S_2 = \bigcup_{(L_{r_i}, L_{r_j}) \in L_{\mathcal{R}} \times L_{\mathcal{R}}} \{\hat{C} \mid C \in \text{MESH}_{h,n_e}(BOTTOM_d^* \cup L_{r_i}, BOTTOM_d^* \cup L_{r_j})\}$$

is also polynomial-sized; furthermore, for any program $P$ in the language $d$-Depth-2-Clause, $P' \in \text{Subclause}_{S_2}$. The second part of the theorem now follows by application of Lemma 9. ∎

An immediate corollary of this result is that Theorems 6 and 5 can be strengthened as follows.

**Corollary 10** *For all constants $d \geq 1$ and $a \geq 2$, the language family*

$$d\text{-Depth-}2\text{-Clause}[\mathcal{DB}, a\text{-}\mathcal{D}et\mathcal{DEC}]$$

*is uniformly polynomially predictable if and only if DNF is polynomially predictable.*

*For all constants $d \geq 1$ and $a \geq 2$, the language family*

$$d\text{-Depth-}2\text{-Clause}'[\mathcal{DB}, a\text{-}\mathcal{D}et\mathcal{DEC}]$$

*is uniformly polynomially predictable if and only if DNF is polynomially predictable.*

---

4. On extended instances of size $n_e$ or less.

Thus in an important sense these learning problems are equivalent to learning boolean DNF. This does not resolve the questions of the learnability of these languages, but does show that their learnability is a difficult formal problem: the predictability of boolean DNF is a long-standing open problem in computational learning theory.

## 5. Related Work

The work described in this paper differs from previous formal work on learning logic programs in simultaneously allowing background knowledge, function-free programs, and recursion. We have also focused exclusively on computational limitations on efficient learnability that are associated with recursion, as we have considered only languages known to be pac-learnable in the nonrecursive case. Since the results of this paper are all negative, we have concentrated on the model of polynomial predictability; negative results in this model immediately imply a negative result in the stronger model of pac-learnability, and also imply negative results for all strictly more expressive languages.

Among the most closely related prior results are the negative results we have previously obtained for certain classes of nonrecursive function-free logic programs (Cohen, 1993b). These results are similar in character to the results described here, but apply to nonrecursive languages. Similar cryptographic results have been obtained by Frazier and Page (1993) for certain classes of programs (both recursive and nonrecursive) that contain function symbols but disallow background knowledge.

Some prior negative results have also been obtained on the learnability of other first-order languages using the proof technique of *consistency hardness* (Pitt & Valiant, 1988). Haussler (1989) showed that the language of "existential conjunction concepts" is not pac-learnable by showing that it can be hard to find a concept in the language consistent with a given set of examples. Similar results have also been obtained for two restricted languages of Horn clauses (Kietz, 1993); a simple description logic (Cohen & Hirsh, 1994); and for the language of sorted first-order terms (Page & Frisch, 1992). All of these results, however, are specific to the model pac-learnability, and none can be easily extended to the polynomial predictability model considered here. The results also do not extend to languages more expressive than these specific constrained languages. Finally, none of these languages allow recursion.

To our knowledge, there are no other negative learnability results for first-order languages. A discussion of prior positive learnability results for first-order languages can be found in the companion paper (Cohen, 1995).

## 6. Summary

This paper and its companion (Cohen, 1995) have considered a large number of different subsets of Datalog. Our aim has been to be not comprehensive, but systematic: in particular, we wished to find precisely where the boundaries of learnability lie as various syntactic restrictions are imposed and relaxed. Since it is all too easy for a reader to "miss the forest for the trees", we will now briefly summarize the results contained in this paper, together with the positive results of the companion paper (Cohen, 1995).

| Local Clauses | Constant-Depth Determinate Clauses | | | | |
|---|---|---|---|---|---|
| $nC_R^-$ | $\boxed{nC_R^-}$ | $nC_R|C_B^-$ | $nC_R, C_B^-$ | $k \times nC_R^-$ | $n \times nC_R^-$ |
| $kC_R^-$ | $\boxed{kC_R^+}$ | $kC_R|C_B^+$ | $kC_R, C_B^{\geq \mathrm{DNF}}$ | $k \times k'C_R^{\geq \mathrm{DNF}}$ | $n \times kC_R^-$ |
| $\boxed{1C_R^-}$ | $\boxed{1C_R^+}$ | $\boxed{1C_R|C_B^+}$ | $\boxed{1C_R, C_B^{=\mathrm{DNF}}}$ | $\boxed{2 \times 1C_R^{=\mathrm{DNF}}}$ | $\boxed{n \times 1C_R^-}$ |

Table 1: A summary of the learnability results

Throughout these papers, we have assumed that a polynomial amount of background knowledge exists; that the programs being learned contain no function symbols; and that literals in the body of a clause have small arity. We have also assumed that recursion is *closed*, meaning that no output variables appear in a recursive clause; however, we believe that this restriction can be relaxed without fundamentally changing the results of the paper.

In the companion paper (Cohen, 1995) we showed that a single nonrecursive constant-depth determinate clause was learnable in the strong model of *identification from equivalence queries*. In this learning model, one is given access to an oracle for counterexamples—that is, an oracle that will find, in unit time, an example on which the current hypothesis is incorrect—and must reconstruct the target program exactly from a polynomial number of these counterexamples. This result implies that a single nonrecursive constant-depth determinate clause is pac-learnable (as the counterexample oracle can be emulated by drawing random examples in the pac setting). The result is not novel (Džeroski et al., 1992); however the proof given is independent, and is also of independent interest. Notably, it is somewhat more rigorous than earlier proofs, and also proves the result directly, rather than via reduction to a propositional learning problem. The proof also introduces a simple version of the *forced simulation* technique, variants of which are used in all of the positive results.

We then showed that the learning algorithm for nonrecursive clauses can be extended to the case of a single *linear recursive* constant-depth determinate clause, leading to the result that this restricted class of recursive programs is also identifiable from equivalence queries. With a bit more effort, this algorithm can be further extended to learn a single $k$-ary recursive constant-depth determinate clause.

We also considered extended the learning algorithm to learn recursive *programs* consisting of more than one constant-depth determinate clauses. The most interesting extension was to simultaneously learn a recursive clause $C_R$ and a base clause $C_B$, using equivalence queries and also a "basecase oracle" that indicates which counterexamples should be covered by the base clause $C_B$. In this model, it is possible to simultaneously learn a recursive clause and a nonrecursive base case in all of the situations for which a recursive clause is learned alone; for instance, one can learn a $k$-ary recursive clause to together with its nonrecursive base case. This was our strongest positive result.

| Language Family | B | R | L/R | Oracles | Notation | Learnable |
|---|---|---|---|---|---|---|
| $d$-DEPTHNONREC$[a\text{-}\mathcal{DB}, a\text{-}\mathcal{D}et\mathcal{DEC}]$ | 1 | 0 | — | EQ | $C_B$ | yes |
| $d$-DEPTHLINREC$[a\text{-}\mathcal{DB}, a\text{-}\mathcal{D}et\mathcal{DEC}]$ | 0 | 1 | 1 | EQ | $1C_R$ | yes |
| $d$-DEPTH-$k$-REC$[a\text{-}\mathcal{DB}, a\text{-}\mathcal{D}et\mathcal{DEC}]$ | 0 | 1 | $k$ | EQ | $kC_R$ | yes |
| $d$-DEPTH-2-CLAUSE$[a\text{-}\mathcal{DB}, a\text{-}\mathcal{D}et\mathcal{DEC}]$ | 1 | 1 | 1 | EQ,BASE | $1C_R\vert C_B$ | yes |
| $kd$-MAXRECLANG$[a\text{-}\mathcal{DB}, a\text{-}\mathcal{D}et\mathcal{DEC}]$ | 1 | 1 | $k$ | EQ,BASE | $kC_R\vert C_B$ | yes |
| $d$-DEPTH-2-CLAUSE$[a\text{-}\mathcal{DB}, a\text{-}\mathcal{D}et\mathcal{DEC}]$ | 1 | 1 | 1 | EQ | $1C_R, C_B$ | =DNF |
| $d$-DEPTH-2-CLAUSE$'[a\text{-}\mathcal{DB}, a\text{-}\mathcal{D}et\mathcal{DEC}]$ | 0 | 2 | 1 | EQ | $2\times 1C_R$ | =DNF |
| $d$-DEPTHLINRECPROG$[a\text{-}\mathcal{DB}, a\text{-}\mathcal{D}et\mathcal{DEC}]$ | 0 | $n$ | 1 | EQ | $n\times 1C_R$ | no |
| $d$-DEPTHREC$[a\text{-}\mathcal{DB}, a\text{-}\mathcal{D}et\mathcal{DEC}]$ | 0 | 1 | $n$ | EQ | $nC_R$ | no |
| $k$-LOCALLINREC$[a\text{-}\mathcal{DB}, a\text{-}\mathcal{DEC}]$ | 0 | 1 | 1 | EQ | $1C_R$ | no |

Table 2: Summary by language of the learnability results. Column B indicates the number of base (nonrecursive) clauses allowed in a program; column R indicates the number of recursive clauses; L/R indicates the number of recursive literals allowed in a single recursive clause; EQ indicates an oracle for equivalence queries and BASE indicates a basecase oracle. For all languages except $k$-LOCALLINREC, all clauses must be determinate and of depth $d$.

These results are summarized in Tables 1 and 2. In Table 1, a program with one $r$-ary recursive clause is denoted $rC_R$, a program with one $r$-ary recursive clause and one nonrecursive basecase is denoted $rC_R, C_B$, or $rC_R\vert C_B$ if there is a "basecase" oracle, and a program with $s$ different $r$-ary recursive clauses is denoted $s\times rC_R$. The boxed results are associated with one or more theorems from this paper, or its companion paper, and the unmarked results are corollaries of other results. A "+" after a program class indicates that it is identifiable from equivalence queries; thus the positive results described above are summarized by the four "+" entries in the lower left-hand corner of the section of the table concerned with constant-depth determinate clauses.

Table 2 presents the same information in a slightly different format, and also relates the notation of Table 1 to the terminology used elsewhere in the paper.

This paper has considered the learnability of the various natural generalizations of the languages shown to be learnable in the companion paper. Consider for the moment single clauses. The companion paper showed that for any fixed $k$ a single $k$-ary recursive constant-depth determinate clause is learnable. Here we showed that all of these restrictions are necessary. In particular, a program of $n$ constant-depth linear recursive clauses is not polynomially predictable; hence the restriction to a single clause is necessary. Also, a single clause with $n$ recursive calls is hard to learn; hence the restriction to $k$-ary recursion is necessary. We also showed that the restriction to constant-depth determinate clauses is necessary, by considering the learnability of *constant locality clauses*. Constant locality clauses are the only known generalization of constant-depth determinate clauses that are pac-learnable in the nonrecursive case. However, we showed that if recursion is allowed, then this language is *not* learnable: even a single linear recursive clause is not polynomially predictable.

Again, these results are summarized in Table 1; a "−" after a program class means that it is not polynomially predictable, under cryptographic assumptions, and hence neither pac-learnable nor identifiable from equivalence queries.

The negative results based on cryptographic hardness give an upper bound on the expressiveness of learnable recursive languages, but still leave open the learnability of programs with a constant number of $k$-ary recursive clauses in the absence of a basecase oracle. In the final section of this paper, we showed that the following problems are, in the model of polynomial predictability, equivalent to predicting boolean DNF:

- predicting two-clause constant-depth determinate recursive programs containing one linear recursive clause and one base case;

- predicting two-clause recursive constant-depth determinate programs containing two linear recursive clauses, even if the base case is known.

We note that these program classes are the very nearly the simplest classes of multi-clause recursive programs that one can imagine, and that the pac-learnability of DNF is a long-standing open problem in computational learning theory. These results suggest, therefore, that pac-learning multi-clause recursive logic programs is difficult; at the very least, they show that finding a provably correct pac-learning algorithm will require substantial advances in computational learning theory. In Table 1, a "= DNF" (respectively ≥ DNF) means that the corresponding language is prediction-equivalent to DNF (respectively at least as hard as DNF).

To further summarize Table 1: with any sort of recursion, only programs containing constant-depth determinate clauses are learnable. The only constant-depth determinate recursive programs that are learnable are those that contain a single $k$-ary recursive clause (in the standard equivalence query model) or a single $k$-ary recursive clause plus a base case (if a "basecase oracle" is allowed). All other classes recursive programs are either cryptographically hard, or as hard as boolean DNF.

## 7. Conclusions

Inductive logic programming is an active area of research, and one broad class of learning problems considered in this area is the class of "automatic logic programming" problems. Prototypical examples of this genre of problems are learning to append two lists, or to multiply two numbers. Most target concepts in automatic logic programming are recursive programs, and often, the training data for the learning system are simply examples of the target concept, together with suitable background knowledge.

The topic of this paper is the pac-learnability of recursive logic programs from random examples and background knowledge; specifically, we wished to establish the computational limitations inherit in performing this task. We began with some positive results established in a companion paper. These results show that one constant-depth determinate closed $k$-ary recursive clause is pac-learnable, and that further, a program consisting of one such recursive clause and one constant-depth determinate nonrecursive clause is also pac-learnable given an additional "basecase oracle".

In this paper we showed that these positive results are not likely to be improved. In particular, we showed that either eliminating the basecase oracle or learning two recursive clauses simultaneously is prediction-equivalent to learning DNF, even in the case of linear recursion. We also showed that the following problems are as hard as breaking (presumably) secure cryptographic codes: pac-learning $n$ linear recursive determinate clauses, pac-learning one $n$-ary recursive determinate clause, or pac-learning one linear recursive $k$-local clause.

These results contribute to machine learning in several ways. From the point of view of computational learning theory, several results are technically interesting. One is the prediction-equivalence of several classes of restricted logic programs and boolean DNF; this result, together with others like it (Cohen, 1993b), reinforces the importance of the learnability problem for DNF. This paper also gives a dramatic example of how adding recursion can have widely differing effects on learnability: while constant-depth determinate clauses remain pac-learnable when linear recursion is added, constant-locality clauses become cryptographically hard.

Our negative results show that systems which apparently learn a larger class of recursive programs must be taking advantage either of some special properties of the target concepts they learn, or of the distribution of examples that they are provided with. We believe that the most likely opportunity for obtaining further positive formal results in this area is to identify and analyze these special properties. For example, in many examples in which FOIL has learned recursive logic programs, it has made use of "complete example sets"—datasets containing all examples of or below a certain size, rather than sets of randomly selected examples (Quinlan & Cameron-Jones, 1993). It is possible that complete datasets allow a more expressive class of programs to be learned than random datasets; in fact, some progress has been recently made toward formalizing this conjecture (De Raedt & Džeroski, 1994).

Finally, and most importantly, this paper has established the boundaries of learnability for determinate recursive programs in the pac-learnability model. In many plausible automatic programming contexts it would be highly desirable to have a system that offered some formal guarantees of correctness. The results of this paper provide upper bounds on what one can hope to achieve with an efficient, formally justified system that learns recursive programs from random examples alone.

## Acknowledgements

The author wishes to thank three anonymous JAIR reviewers for a number of useful suggestions on the presentation and technical content.